\documentclass[preprint,12pt]{elsarticle}

\usepackage{amsmath,amssymb}
\usepackage{graphicx}
\usepackage{booktabs}
\usepackage{multirow}
\usepackage{array}
\usepackage{longtable}
\usepackage[table]{xcolor}
\usepackage{url}
\usepackage{subcaption}
\usepackage{placeins}
\usepackage[colorlinks=true,linkcolor=blue,citecolor=blue,urlcolor=blue]{hyperref}

\journal{Process Safety and Environmental Protection}

\begin{document}

\begin{frontmatter}

\title{Evaluating Near-Perfect Accuracy Claims in Machine-Learning-Based
Leak Detection for Two-Phase Pipelines: A Condition-Blocked Audit}

\author[mun]{Issah N. Suleiman}
\author[mun]{Sormeh Serpoosh}
\author[hbku]{Nadine Elkholy}
\author[boumerdes]{Hicham Ferroudji}
\author[hbku]{Mohammad Azizur Rahman}
\author[mun]{Matthew Hamilton\corref{cor1}}
\ead{mhamilton@mun.ca}
\cortext[cor1]{Corresponding author}
\address[mun]{Memorial University of Newfoundland, St.~John's,
Newfoundland and Labrador, Canada}
\address[hbku]{College of Science and Engineering, Hamad Bin Khalifa
University, Education City, Doha, Qatar}
\address[boumerdes]{Laboratory of Petroleum Equipment's Reliability and
Materials, Faculty of Hydrocarbons and Chemistry, University M'Hamed
Bougara of Boumerdes, Boumerdes, Algeria}

\begin{abstract}
Reliable leak detection in gas--liquid two-phase pipelines is important for process safety, yet machine-learning studies frequently report near-perfect performance using evaluation protocols that may not represent deployment to unseen operating conditions. This study audits a pressure-based leak-detection model using condition-blocked leave-one-condition-out validation, in which the classifier is tested on a flow condition excluded from training. Accuracy decreased from 93.3\% under within-condition evaluation to 59.1\% across fifteen unseen-condition tests. Ablation analysis showed that most of this difference arose from the data-splitting protocol, with a smaller contribution from encoding operating conditions as model inputs. Under condition-blocked evaluation, the false-alarm and missed-detection rates increased to 34.1\% and 10.2\%, respectively, demonstrating that near-perfect reported accuracy did not translate into reliable operation at unseen flow conditions. The reduction persisted across all evaluated classifiers, indicating that it was caused by the evaluation design rather than a particular model architecture. Grouped evaluation of an independent acoustic pipeline dataset produced a comparable reduction toward chance-level performance. These findings show that validation units must represent the conditions encountered during deployment. Condition-blocked evaluation and explicit reporting of operational error rates are therefore necessary before machine-learning leak detectors can support process-safety decisions.
\end{abstract}

\begin{keyword}
condition-blocked validation \sep evaluation bias \sep multiphase flow \sep
leak detection \sep false alarm rate \sep process safety
\end{keyword}

\end{frontmatter}

\section*{Nomenclature}
\begin{tabular}{ll}
$W_k$                     & Sliding window size (samples)                       \\
$V_{SL}$                  & Liquid superficial velocity (m/s)                   \\
$V_{SG}$                  & Gas superficial velocity (m/s)                      \\
$K$                       & Number of LOCO cross-validation folds               \\
$\Delta_{\text{split}}$   & Split-protocol accuracy gap (percentage points)     \\
$\Delta_{\text{enc}}$     & Feature-encoding accuracy gap (percentage points)   \\
$I$                       & Interaction contrast (percentage points)            \\
FAR                       & False alarm rate                                    \\
MDR                       & Missed detection rate                               \\
FSO                       & Full-scale output                                   \\
LOCO                      & Leave-One-Condition-Out cross-validation            \\
C0, C1, C2                & No-leak, single-point, two-point leakage classes    \\
S1, S2                    & Upstream and downstream pressure sensors            \\
P1, P2                    & Feature vectors from S1 and S2 respectively         \\
MC4                       & Primary cascaded CatBoost model                     \\
RMS                       & Root mean square                                    \\
PSEP                      & Process Safety and Environmental Protection         \\
\end{tabular}

\section{Introduction}
\label{sec:intro}

Leak detection in gas--liquid two-phase pipelines is not an academic
benchmark problem; it is a process-safety requirement. Undetected leakage
events carry consequences ranging from environmental contamination to
facility loss, and alarm systems that generate excessive false positives
erode operator confidence to the point where genuine events are dismissed.
A classifier that reports 100\% accuracy during development and then
triggers one in three alarms falsely during field operation is not a
marginal failure --- it is a systemic one. The distinction between
within-condition performance, where the model is tested on conditions it
has already seen during training, and unseen-condition generalisation, where
it is not, is therefore a process-safety question, not merely a machine
learning benchmarking question. That distinction is what this paper
addresses.

Machine-learning classifiers applied to dynamic pressure signals from
two-phase pipelines have recently reached or approached 100\% reported
classification accuracy~\cite{alquennah2026,ferroudji2025}. The study
subjected to formal evaluation here~\cite{alquennah2026} was conducted
within the same research programme as this paper, using the same six-metre
horizontal air--water flow loop. Re-evaluating the published claims on
the same data and apparatus, under a protocol that requires the model to
classify windows from a flow condition it was not trained on, reduces
accuracy from 93.3\%$\pm$0.7\% (within-condition) to 59.1\%$\pm$0.6\%
on the full 15-condition dataset and to 50.9\%$\pm$0.5\% on the nine
conditions matching the original study design. The false alarm rate under
this condition-blocked protocol rises from 0.0\% to 34.1\%.

The pattern is not confined to this apparatus or to leak detection. In ship fuel-consumption modelling using 1~Hz operational data, Chittamuru et al.~\cite{chittamuru2026} likewise showed that random train--test splitting admits temporal leakage and yields optimistic performance relative to time-aware validation. In wind-speed forecasting,
an inappropriate cross-validation design inflated reported performance by
20.5\% compared with a leakage-free protocol~\cite{sahin2026}. In
repeated-measures biomechanical studies, switching to participant-level
cross-validation reduced reported performance from 0.91 to
0.66~\cite{karbalaie2026}. In geospatial modelling, random train--test
splitting overestimates performance by up to 0.16~AUC and 80\% relative
MAE relative to spatially blocked evaluation~\cite{spatialcv}.
Bergmeir et al.~\cite{bergmeir2018} established the theoretical conditions
under which cross-validation is and is not valid for autocorrelated time
series, underpinning what practitioners have documented empirically across
domains. Feature-engineering guidelines have long cautioned that any
transformation that sees outside the training fold will produce optimistic
reported performance~\cite{featureeng}, and a systematic review of ML
reporting standards found that fewer than one-third of published studies
include evaluation-validity elements~\cite{klement2023}.

No prior study in the multiphase pipeline leak-detection literature has
applied a quantitative decomposition of this kind, nor isolated each source
of accuracy inflation rather than treating data leakage as a single
phenomenon. This paper makes five contributions.

\begin{enumerate}
\item A \textbf{leakage decomposition framework} that quantifies two
accuracy-gap components --- the split-protocol component and the
operating-condition feature-encoding component --- through a complete
$2\times2$ four-condition ablation (Section~\ref{sec:decomp}).

\item Empirical evidence that the two components are \textbf{approximately
additive}: an interaction contrast $I$, measuring the non-additive remainder
after combining the split-protocol and feature-encoding effects, remains within
$\pm 3$~pp at all three window sizes, compared with the approximately 34~pp
total gap.

\item Documentation of a previously unreported \textbf{dual-sensor coverage
structure}: the downstream sensor (S2) carries explicit metadata labels
for nine of fifteen operating conditions; the remaining six are treated as
inferred and confirmed non-duplicate by pairwise correlation analysis.

\item A \textbf{corrected bidirectional noise robustness protocol} in which
training and test data receive identical synthetic noise, reflecting the
field deployment scenario more accurately than the asymmetric approach in
the original study.

\item \textbf{External evaluation} on the independent GPLA-12 acoustic
gas-pipeline dataset~\cite{li2021gpla}, showing the pattern holds across
a different physical system and sensor modality
(Section~\ref{sec:external}).
\end{enumerate}

Section~\ref{sec:related} reviews related work. Section~\ref{sec:methods}
describes the dataset, the evaluation protocol, and the corrected
methodology. Section~\ref{sec:results} reports the corrected results.
Section~\ref{sec:discussion} discusses engineering and safety implications.
Section~\ref{sec:conclusion} concludes.

\section{Related Work}
\label{sec:related}

\subsection{Evaluation Validity in Time-Series and Grouped Experiments}

Bergmeir et al.~\cite{bergmeir2018} established that cross-validation is
valid for time-series prediction only when specific conditions are met ---
conditions violated when windows constructed from the same recording appear
in both training and test sets. \c{S}ahin et al.~\cite{sahin2026}
quantified the empirical consequence in wind-speed forecasting: a 20.5\%
accuracy gap between an appropriate and an inappropriate cross-validation
design. In maritime fuel-consumption modelling, Chittamuru et al.~\cite{chittamuru2026} applied time-aware validation to 1~Hz operational data from the CCGS \textit{Sir Wilfrid Laurier}, demonstrating that random splitting of strongly autocorrelated records can overstate deployment-relevant performance and alter model rankings. Feature-engineering practice for time-series ML has documented
the same risk for decades~\cite{featureeng}, but the guideline is not
uniformly applied in domain-specific literature, including the pipeline
monitoring field.

\subsection{The Grouping-Unit Problem}

Spatial autocorrelation causes random train--test splits to overstate
geospatial model performance by up to 0.16~AUC and 80\% relative MAE
compared with spatially blocked evaluation~\cite{spatialcv}. In
repeated-measures biomechanics, participant-level cross-validation reduced
reported model performance from 0.91 to 0.66 relative to random
window-level splits~\cite{karbalaie2026}. In the present experimental
setting, two pressure recordings at the same gas--liquid superficial-velocity pair $(V_{SG}, V_{SL})$
--- taken at different times during the same session --- share far more
statistical structure than two recordings at different conditions taken
seconds apart. The independence unit is the condition, not the window.
Leave-one-condition-out cross-validation implements this principle
for small factorial experimental designs; it is not proposed here as
a novel algorithm.

\subsection{Reporting Standards}

Klement and El~Emam~\cite{klement2023} reviewed ML reporting guidelines
across medical research and found that evaluation-validity elements appear
in fewer than one-third of published studies. The present paper addresses
this gap by providing a quantifiable decomposition protocol with each
component individually estimated and its interaction assessed empirically.

\subsection{ML for Multiphase Pipeline Leak Detection}

Representative contributions to ML-based leak detection in two-phase
pipelines, spanning experimental, numerical, and data-driven approaches,
are summarised in Table~\ref{tab:literature}. The study audited
here~\cite{alquennah2026} uses the same apparatus as related work from
the same programme: Ferroudji et al.~\cite{ferroudji2025} characterised
leak impacts using statistical and wavelet-transform methods;
\cite{ferroudji2026} extended that work to multi-leak comparison;
\cite{ferroudji2024b} provided the flow-modelling foundation; and
Barooah et al.~\cite{barooah2024} first applied time-series ML to the
same configuration.

\section{Methodology}
\label{sec:methods}

\subsection{Experimental Apparatus}
\label{sec:apparatus}

\begin{figure}[htbp]
  \centering
  \includegraphics[width=\linewidth]{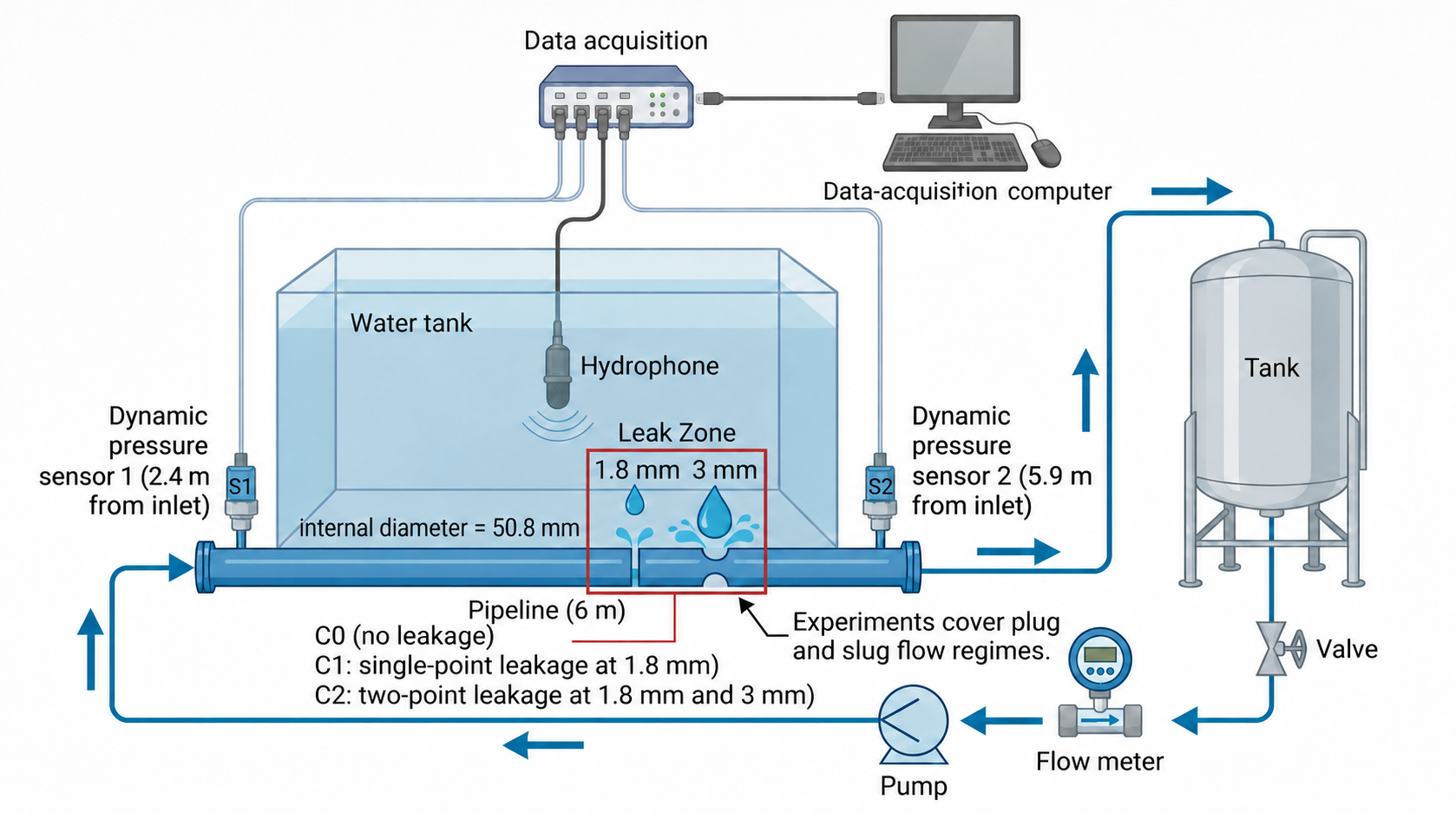}
  \caption{Schematic of the experimental two-phase flow loop
  (50.8~mm ID, 6~m): pressure sensors S1 (2.4~m from inlet) and
  S2 (5.9~m from inlet), and the two-orifice leak zone
  (1.8~mm and 3~mm) at 3.85~m. Dynamic pressure sampled at 2~Hz.
  Only S1 and S2 pressure signals were used in this study.}
  \label{fig:schematic}
\end{figure}

The experimental setup is shown in Fig.~\ref{fig:schematic}. The apparatus
used in~\cite{alquennah2026} is a closed-loop horizontal air--water
two-phase flow pipeline with internal diameter 50.8~mm and total length
6~m. A leak zone containing two orifice ports (diameters 1.8~mm and 3~mm)
is located 3.85~m downstream of the inlet. Three leakage conditions are
studied: no leakage (C0), activation of the 1.8~mm orifice only (C1,
single-point leakage), and simultaneous activation of both orifices (C2,
two-point or multiple leakage, following the terminology
of~\cite{alquennah2026}). Flow regimes across the experimental conditions
span plug and slug patterns.

Dynamic pressure was recorded using two piezoelectric transducers: sensor
S1 at 2.4~m upstream of the leak zone and sensor S2 at 5.9~m downstream.
Both sampled at 2~Hz. Each experimental condition was held at steady state
for 30~s per recording.

\begin{figure}[htbp]
  \centering
  \includegraphics[width=\linewidth]{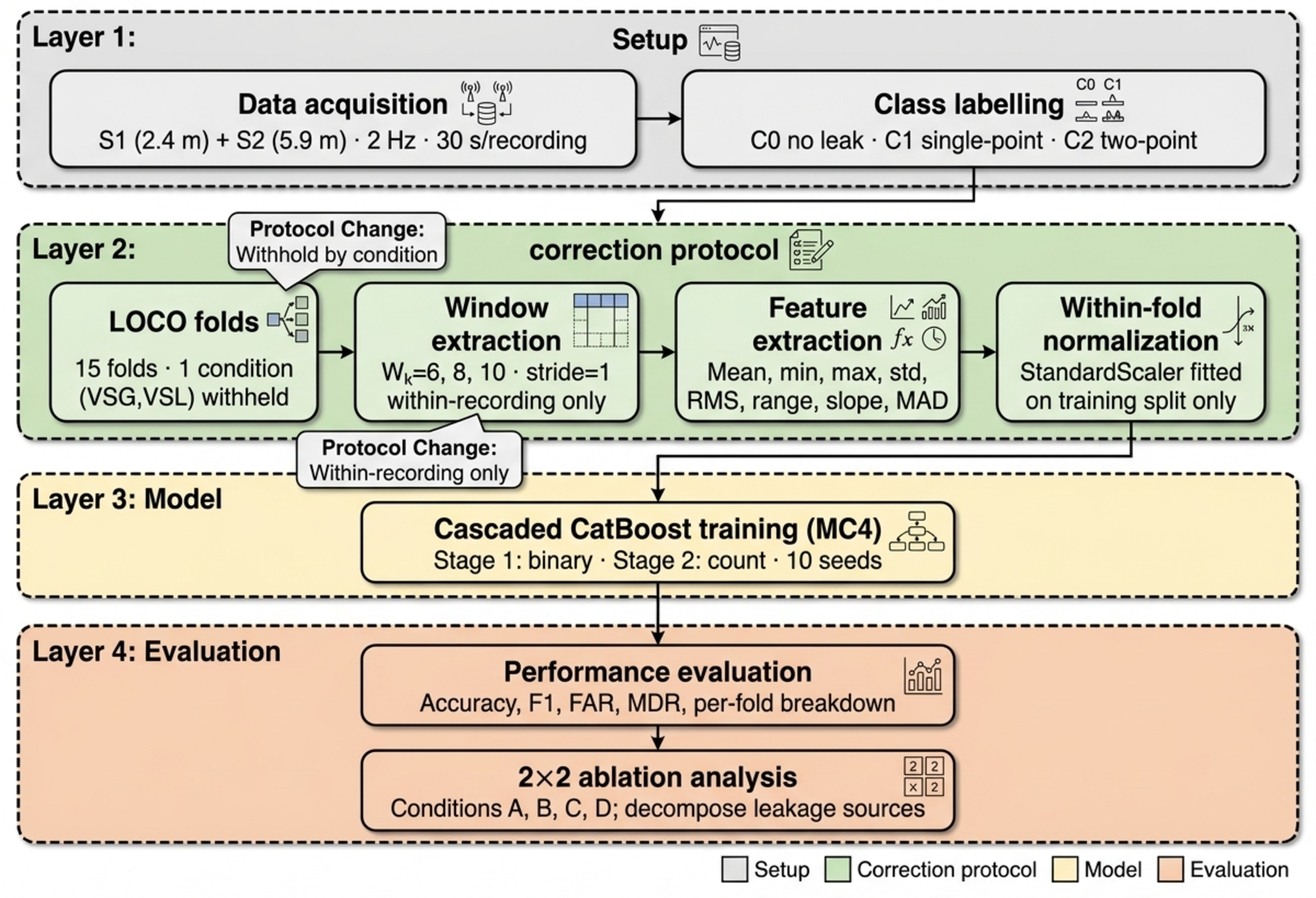}
  \caption{Corrected ML audit pipeline for multiphase leak classification
  (C0: no leak; C1: single-point; C2: two-point/multiple leakage). Two
  protocol changes are enforced relative to~\cite{alquennah2026}: windows
  are extracted within individual 30-second recordings only, and LOCO
  withholds each $(V_{SG}, V_{SL})$ condition from training in turn.
  \texttt{StandardScaler} is fitted on the training split of each fold
  only.}
  \label{fig:flowchart_linear}
\end{figure}

\begin{figure}[htbp]
  \centering
  \includegraphics[width=\linewidth]{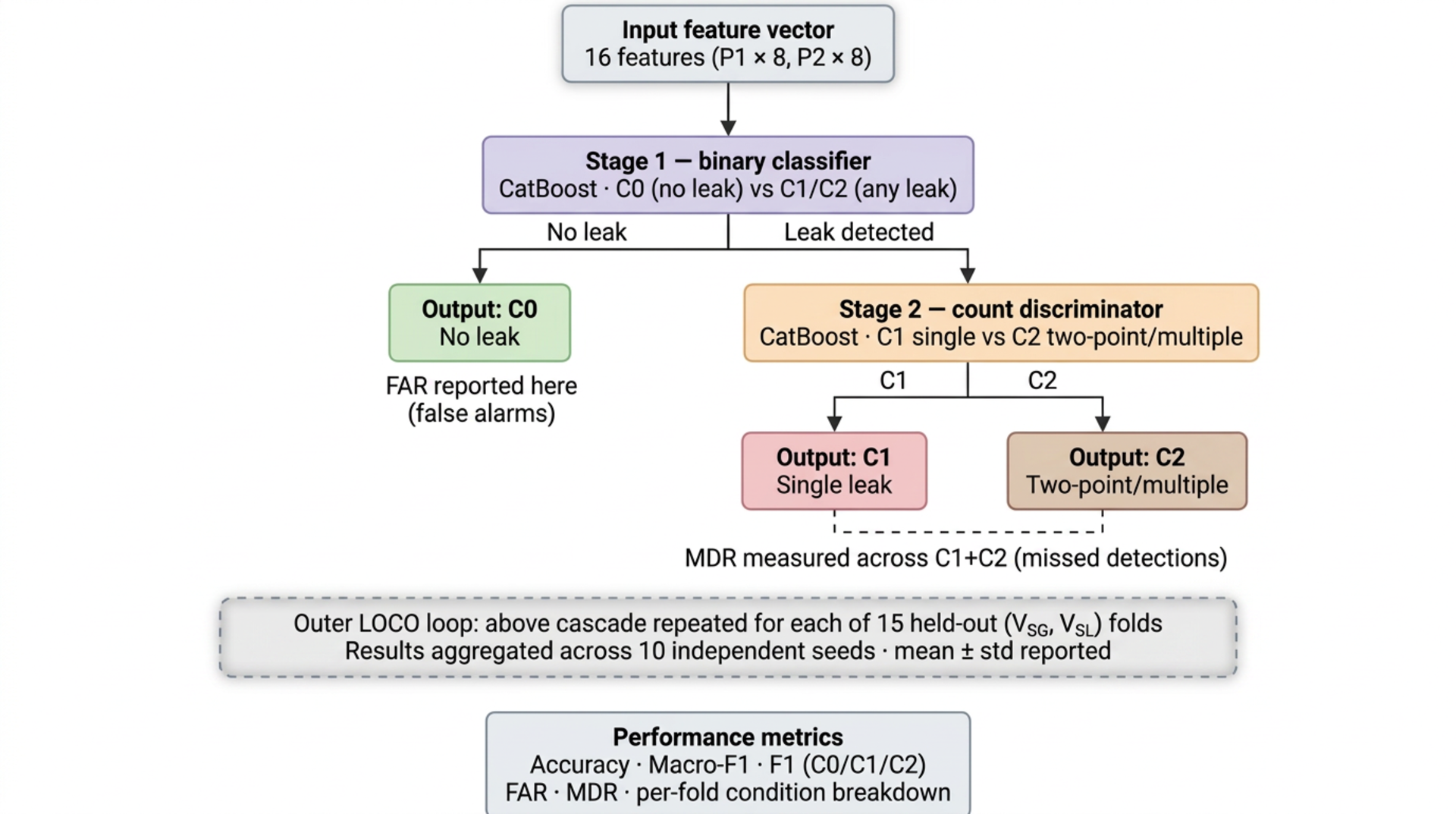}
  \caption{Two-stage cascaded classifier architecture (MC4). Stage~1
  performs binary detection (C0 versus C1 or C2); windows classified as
  leaking are forwarded to Stage~2, which discriminates single-point (C1)
  from two-point/multiple leakage (C2). FAR is measured at Stage~1; MDR
  is measured across Stage~2. The LOCO loop repeats both stages for each
  of the fifteen held-out conditions, aggregating results across ten
  independently seeded trials.}
  \label{fig:flowchart_cascade}
\end{figure}

\subsection{Dataset Structure and the Data Hierarchy}
\label{sec:dataset}

The term ``sample'' refers to different entities at different levels of
the data hierarchy, and these must be distinguished precisely.

\begin{itemize}
  \item \textbf{Operating condition:} A unique $(V_{SG}, V_{SL})$
  combination. Fifteen conditions were defined---three gas velocities
  ($V_{SG}\in\{0.063, 0.12, 0.19\}$~m/s) and five liquid velocities
  ($V_{SL}\in\{1.03, 1.34, 1.63, 1.94, 2.21\}$~m/s).
  \item \textbf{Recording (trace):} One 30-second continuous pressure
  time series at a given condition and leakage class. There is one trace
  per (condition, class) cell --- 45 traces per sensor in total.
  \item \textbf{Raw observation:} One pressure reading at 2~Hz. Each
  30-second trace yields 61 raw observations ($30 \times 2 + 1$).
  \item \textbf{Constructed window:} A sliding window of $W_k$ consecutive
  raw observations from a single trace. At $W_k=10$, stride~1, one trace
  yields $61 - 10 + 1 = 52$ windows. Windows from the same trace are
  correlated pseudoreplicates, not independent experimental runs.
\end{itemize}

The count of 1,647 in Alquennah et al.~\cite{alquennah2026} refers to
raw pressure observations: $9$ operating conditions $\times$ $3$ leakage
classes $\times$ $61$ raw observations per condition--class recording
$= 1{,}647$ observations per sensor. That study used the nine conditions
carrying explicit S2 metadata labels ($V_{SL}\in\{1.03, 1.63, 2.21\}$~m/s).
The full 15-condition dataset analysed here contains $15 \times 3 \times
61 = 2{,}745$ raw observations across both sensors. All sample-count
references in this paper use ``raw observations'' for the 2-Hz pressure
readings and ``windows'' for the constructed feature instances.

\textbf{Sensor coverage.} The S2 source files carry explicit VSL metadata
labels for nine of fifteen conditions. The remaining six pressure-signal series lacked operating-condition labels but contained valid numerical measurements. These unlabelled series are treated as inferred operating conditions based on pairwise Pearson correlation analysis within each $V_{SG}$ group
(maximum observed correlation $r = 0.837$, far below the $r > 0.999$
threshold that would indicate duplication) and monotonically increasing
column means consistent with the pressure--velocity relationship in
two-phase flow. Because the original operating-condition metadata were unavailable, the six inferred assignments could not be independently verified. Results based on all 15 conditions are therefore reported alongside the nine-condition analysis as a sensitivity assessment. Alquennah et al.~\cite{alquennah2026} used the
nine explicitly labelled conditions (POOL\textsubscript{9}); the full
15-condition pool (POOL\textsubscript{15}) forms the primary analysis here.

\subsection{Evaluation Protocol Characteristics in the Original Study}
\label{sec:issues}

Re-examining the original data-processing pipeline identified four protocol
characteristics that contribute to the reported accuracy.

\begin{enumerate}
\item \textbf{Within-condition evaluation.}
The published protocol uses an 80/20 time-based split that retains all
operating conditions in both training and test sets. This measures
within-condition performance: can the model classify later recordings from
a condition it has already seen? It does not measure transfer to an
entirely new condition. The present study compares this estimand with a
condition-blocked LOCO estimand designed to measure that transfer.

\item \textbf{Operating-condition feature encoding.}
Ground-truth $V_{SL}$ and $V_{SG}$ values were included as direct model
inputs. Including these as features is not inherently inappropriate: if
the deployed system has access to flow-meter readings or set-point data,
they are legitimate covariates. What is tested here is whether performance
is maintained when those features are withheld --- a relevant question
for deployment scenarios where the precise operating condition is unknown.
This is properly described as a feature-ablation test.

\item \textbf{Asymmetric noise evaluation.}
Noise robustness was assessed by training on clean data and testing on
synthetically contaminated data. We replace this with symmetric
contamination of both training and test, reflecting the scenario where
historical sensor recordings and live inference inputs share the same
noise characteristics.

\item \textbf{Sensor-coverage structure.}
S2 carries explicit VSL labels for nine of fifteen conditions, as described
above. The present analysis includes all fifteen.
\end{enumerate}

\subsection{Corrected Machine Learning Pipeline}
\label{sec:mlpipeline}

The corrected pipeline (Fig.~\ref{fig:flowchart_linear}) and the
two-stage cascade architecture (Fig.~\ref{fig:flowchart_cascade}) follow
the structure of~\cite{alquennah2026} with the protocol modifications
enumerated above.

\subsection{Feature Extraction}
\label{sec:features}

Eight statistical features are extracted per sensor per window: mean,
minimum, maximum, standard deviation, RMS, range, linear-fit slope, and
mean absolute difference. These match the features used in~\cite{alquennah2026},
allowing direct methodological comparison. Feature vectors are standardised
using a \texttt{StandardScaler} fitted exclusively on the training windows
of each LOCO fold; the same parameters are applied to validation and test
windows without refitting, so no statistical information from the held-out
condition enters the preprocessing step.

\subsection{Leave-One-Condition-Out Protocol}
\label{sec:loco}

In each of the fifteen LOCO folds, all windows associated with one
$(V_{SG}, V_{SL})$ condition are withheld as the test set; the remaining
fourteen form the training set. Within the training fold, an 80/20
stratified random partition produces inner training and validation subsets
for hyperparameter selection and early stopping. This inner split operates
at the window level, which allows adjacent overlapping windows from the
same recording to appear in both inner subsets, repeating in miniature the
within-condition dependence documented as Issue~1. The hyperparameter
sensitivity analysis (Section~\ref{sec:hpsensitivity}) shows a modest
practical impact (approximately 2.3~pp at the available sample sizes), but
the tuning results should be treated as potentially optimistic. A
recording-block or condition-grouped inner split would eliminate this
dependence and is recommended for future studies.

\subsection{Decomposition Protocol}
\label{sec:decomp}

The split-protocol and feature-encoding contributions are isolated through
a complete $2\times2$ ablation. Each of the four experimental conditions
crosses the two protocol dimensions --- split type (within-condition
versus condition-blocked) and feature set (with versus without $V_{SL}$
and $V_{SG}$):

\begin{enumerate}
\item \textbf{Condition A --- published protocol.} Time-based 80/20 split;
$V_{SL}$ and $V_{SG}$ included as inputs. Reproduces the
estimand of~\cite{alquennah2026}.
\item \textbf{Condition B --- split corrected.} LOCO applied; $V_{SL}$
and $V_{SG}$ retained as inputs.
\item \textbf{Condition C --- both corrected (primary result).} LOCO
applied; $V_{SL}$ and $V_{SG}$ excluded.
\item \textbf{Condition D --- feature set corrected only.} Time-based
split retained; $V_{SL}$ and $V_{SG}$ excluded. Without this cell the
two components are order-dependent and cannot be assessed separately.
\end{enumerate}

The two accuracy-gap components are defined as:
\begin{align}
\Delta_{\text{split}} &= \text{Acc}(A) - \text{Acc}(B), \label{eq:split}\\
\Delta_{\text{enc}}   &= \text{Acc}(B) - \text{Acc}(C). \label{eq:enc}
\end{align}

Approximate additivity is assessed via the interaction contrast
\begin{equation}
I = \text{Acc}(A) - \text{Acc}(B) - \text{Acc}(D) + \text{Acc}(C),
\label{eq:interact}
\end{equation}
which equals zero when the two components combine without interaction. A
small $|I|$ relative to $\Delta_{\text{split}}$ and $\Delta_{\text{enc}}$
supports approximate additivity; a large $|I|$ signals a substantial
interaction between the two corrections.

Each experimental condition is evaluated with ten independently drawn
random seeds. The primary model, MC4 (cascaded CatBoost), uses the fixed
configuration \texttt{iterations}~=~300, \texttt{depth}~=~6,
\texttt{learning\_rate}~=~0.05. Hyperparameter sensitivity is quantified
separately against a nested grid search over 18 combinations
($\{200,300\}\times\{4,6,8\}\times\{0.03,0.05,0.10\}$). Standard
deviations across seeds quantify model-training instability --- sensitivity
to random initialisation within a fixed data partition --- rather than
uncertainty across independently repeated physical experiments.

\subsection{Models Evaluated}
\label{sec:models}

The primary model is MC4 (Fig.~\ref{fig:flowchart_cascade}), introduced
in~\cite{alquennah2026}. For comparison, the seven single-stage classifiers
from the original study are also evaluated under the corrected protocol:
$k$-nearest neighbours (k-NN), support vector machine (SVM), decision tree
(DT), random forest (RF), single-stage CatBoost, max-voting ensemble, and
stacking ensemble. A feed-forward neural network (256-128-64 units, batch
normalisation, dropout rate 0.30, Adam optimiser, patience~15) is included
as an additional baseline.

\subsection{Corrected Noise Robustness Protocol}
\label{sec:noise}

Synthetic Gaussian noise is added to both training and test feature vectors
at identical magnitude, expressed as a percentage of FSO. FSO is derived
per feature from the training-fold range with a 10\% margin; the noise
level is treated as a $3\sigma$ interval, and all FSO and scaling
parameters are estimated from the training fold only. Five levels are
evaluated (0.0, 0.1, 0.2, 0.5, 1.0\% FSO) under LOCO. Symmetric
contamination represents one plausible deployment scenario --- it models
the case where historical sensor recordings and live inference inputs share
the same noise floor. Other scenarios are equally plausible and are
not evaluated here.

\subsection{Performance Metrics}
\label{sec:metrics}

Overall accuracy and macro-averaged F1-score are the primary aggregate
metrics. Per-class F1-score is reported for each leakage class. FAR is
the proportion of C0 windows classified as leaking; MDR is the proportion
of C1 or C2 windows classified as no-leak. All metrics are mean~$\pm$
standard deviation across ten seeds, where the standard deviation reflects
model-training instability, not uncertainty from independently repeated
physical experiments.

\section{Results}
\label{sec:results}

\subsection{Complete $2\times2$ Ablation}
\label{sec:ablation}

\begin{figure}[htbp]
  \centering
  \includegraphics[width=\linewidth]{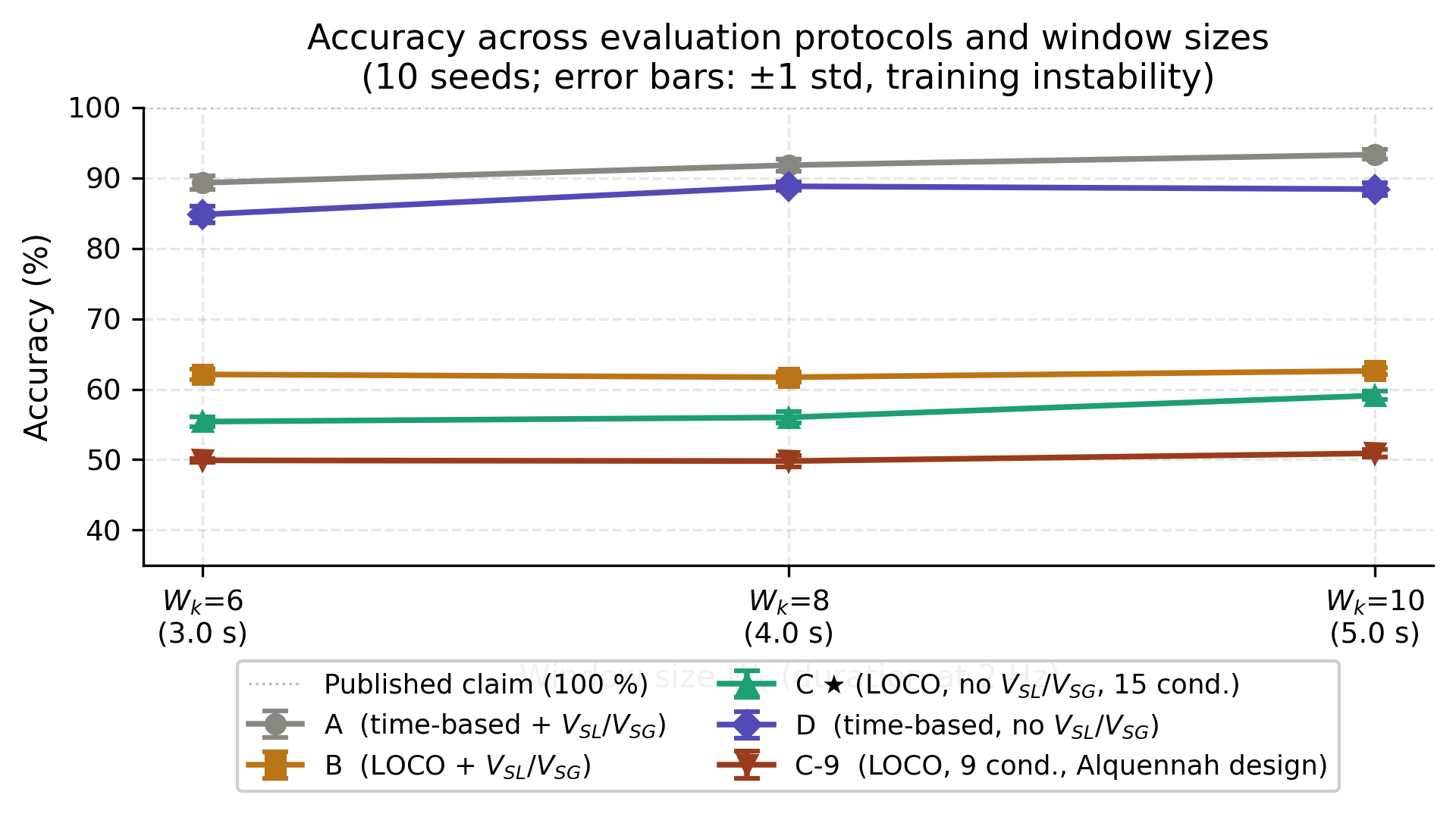}
  \caption{Accuracy across evaluation protocols at three window sizes
  ($W_k = 6$, 8, 10). Error bars: $\pm1$ standard deviation across ten
  seeds, reflecting model-training instability. The dashed line marks
  the accuracy reported in~\cite{alquennah2026}. Condition A uses a
  time-based 80/20 split retaining all conditions in training and test.}
  \label{fig:fig1}
\end{figure}

Table~\ref{tab:table1} and Fig.~\ref{fig:fig1} summarise accuracy across
all four conditions and three window sizes. At $W_k=10$, the within-condition
accuracy of 93.3\%$\pm$0.7\% (Condition~A) falls to 59.1\%$\pm$0.6\% under
full correction (Condition~C, POOL\textsubscript{15}) --- a 34.2-percentage-point
gap. On the nine-condition sensitivity run matching the original study
design (POOL\textsubscript{9}), corrected accuracy is 50.9\%$\pm$0.5\%.

\FloatBarrier
\begin{table}[htbp]
\caption{Complete $2\times2$ ablation: accuracy (mean $\pm$ std across
10 seeds). The interaction contrast $I = A - B - D + C$ quantifies
approximate additivity; values $|I|\leq3$~pp at all window sizes indicate
the two components combine with negligible interaction.}
\label{tab:table1}
\centering
\small
\begin{tabular}{lccc}
\toprule
Protocol & $W_k$=6 & $W_k$=8 & $W_k$=10 \\
\midrule
\textbf{A}: Time-based + $V_{SL}$/$V_{SG}$
  & $89.3\pm1.0$\% & $91.8\pm0.9$\% & $93.3\pm0.7$\% \\
\textbf{B}: LOCO + $V_{SL}$/$V_{SG}$
  & $62.1\pm0.7$\% & $61.7\pm0.7$\% & $62.6\pm0.5$\% \\
\textbf{C}$^{\star}$: LOCO, no $V_{SL}$/$V_{SG}$ (15 cond.)
  & $55.4\pm0.7$\% & $56.0\pm0.8$\% & $59.1\pm0.6$\% \\
\textbf{D}: Time-based, no $V_{SL}$/$V_{SG}$
  & $84.8\pm1.2$\% & $88.8\pm0.6$\% & $88.4\pm0.9$\% \\
C-9 sensitivity (9 conditions)
  & $49.9\pm0.3$\% & $49.8\pm0.8$\% & $50.9\pm0.5$\% \\
\midrule
$\Delta_{\text{split}}$ (A$\to$B)
  & $-27.2$\,pp & $-30.1$\,pp & $-30.7$\,pp \\
$\Delta_{\text{enc}}$ (B$\to$C)
  & $-6.7$\,pp  & $-5.7$\,pp  & $-3.5$\,pp  \\
Total (A$\to$C)
  & $-33.9$\,pp & $-35.8$\,pp & $-34.2$\,pp \\
$\Delta_{\text{enc}}$ alone (A$\to$D)
  & $-4.5$\,pp  & $-3.0$\,pp  & $-4.9$\,pp  \\
$\Delta_{\text{split}}$ alone (D$\to$C)
  & $-29.5$\,pp & $-32.8$\,pp & $-29.3$\,pp \\
Interaction contrast $I$
  & $-2.2$\,pp  & $-2.7$\,pp  & $+1.4$\,pp  \\
\bottomrule
\end{tabular}\\[4pt]
\footnotesize $^{\star}$Primary corrected result.
Interaction contrast $I = A - B - D + C$ (Eq.~\ref{eq:interact}):
all $|I| \leq 3$~pp, supporting approximate additivity of the two
components.
\end{table}

The interaction contrast (Eq.~\ref{eq:interact}) is $-2.2$~pp at $W_k=6$,
$-2.7$~pp at $W_k=8$, and $+1.4$~pp at $W_k=10$. All values are small
relative to the dominant split-protocol component of 27--31~pp, supporting
approximate additivity of the two corrections. The split-protocol component
accounts for approximately 90\% of the total gap at $W_k=10$. The
feature-encoding component is larger at shorter window sizes (6.7~pp at
$W_k=6$ versus 3.5~pp at $W_k=10$), consistent with shorter windows
encoding less condition information in their statistical summaries and
therefore deriving more discriminative gain from the explicit
$V_{SL}$/$V_{SG}$ features.

\subsection{Nine-Condition Sensitivity Analysis}

The POOL\textsubscript{9} run uses the nine conditions with explicit S2
labels, matching the scope of~\cite{alquennah2026} exactly. At $W_k=10$,
corrected accuracy is 50.9\%$\pm$0.5\% --- approximately 49 percentage
points below the reported 100.0\%. The same conditions, the same 1,647
raw observations, the same sensor configuration: the only difference is
whether the model encounters the test condition during training.

\subsection{Hyperparameter Sensitivity}
\label{sec:hpsensitivity}

Comparing fixed default hyperparameters against per-fold nested grid
search (18 combinations) at $W_k=10$ gives a 2.3-point accuracy
difference, within one standard deviation of either configuration. FAR
and MDR are essentially unchanged. The primary result is not sensitive
to the hyperparameter selection strategy at the available sample sizes,
though the window-level inner split introduces a residual dependence
discussed in Section~\ref{sec:loco}.

\subsection{Per-Class Results at $W_k=10$, Condition C}
\label{sec:perclass}

Table~\ref{tab:table2} reports per-class F1-score averaged across ten seeds.
The no-leak class (C0) achieves the highest F1 (70.8\%); two-point/multiple
leakage (C2) achieves the lowest (50.3\%).

\begin{table}[htbp]
\caption{Per-class metrics: $W_k=10$, LOCO, no $V_{SL}$/$V_{SG}$, 15
conditions, 10 seeds. FAR and MDR are laboratory estimates from a single
rig; they should not be used as deployment readiness criteria without a
prospectively designed field study at the intended operating conditions.}
\label{tab:table2}
\centering
\begin{tabular}{lcc}
\toprule
Class & F1 (\%) & Function \\
\midrule
C0: No leak          & $70.8\pm0.8$ & Binary alarm         \\
C1: Single-point     & $56.6\pm0.8$ & Count discrimination \\
C2: Two-point/mult.  & $50.3\pm0.7$ & Count discrimination \\
\midrule
Macro F1             & $59.2\pm0.6$ &                      \\
Accuracy             & $59.1\pm0.6$ &                      \\
FAR (laboratory)     & $34.1\pm1.0$ & C0 classified as leak \\
MDR (laboratory)     & $10.2\pm0.4$ & C1/C2 classified as C0 \\
\bottomrule
\end{tabular}
\end{table}

\begin{figure}[htbp]
  \centering
  \includegraphics[width=\linewidth]{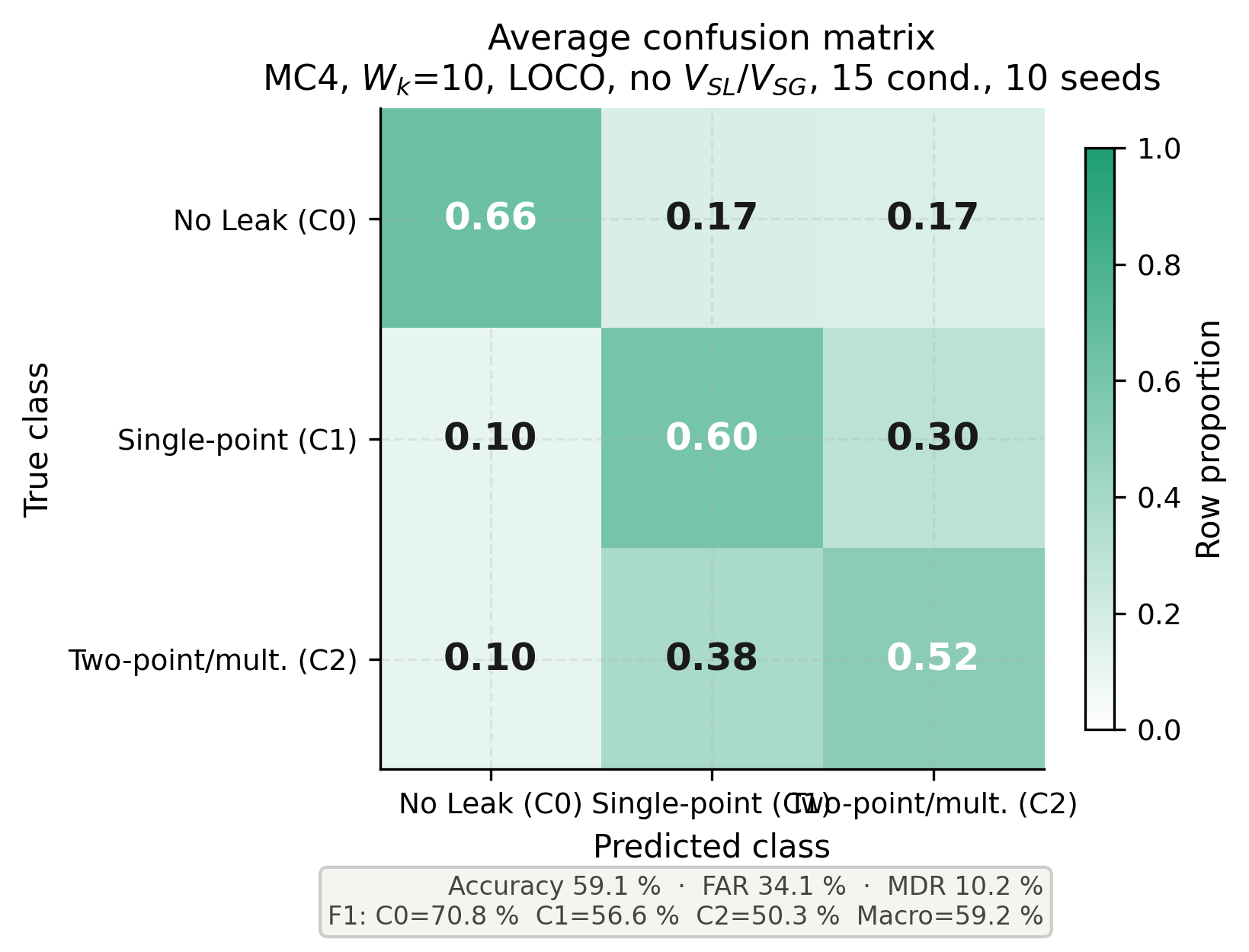}
  \caption{Normalised confusion matrix at $W_k=10$, LOCO, no
  $V_{SL}$/$V_{SG}$, 15 conditions, averaged over ten seeds. Class C2
  is two-point or multiple leakage, consistent with~\cite{alquennah2026}.
  Overall accuracy: 59.1\%; FAR: 34.1\%; MDR: 10.2\%.}
  \label{fig:fig2}
\end{figure}

\subsection{All Models Under the Condition-Blocked Protocol}
\label{sec:allmodels}

\begin{figure}[htbp]
  \centering
  \includegraphics[width=\linewidth]{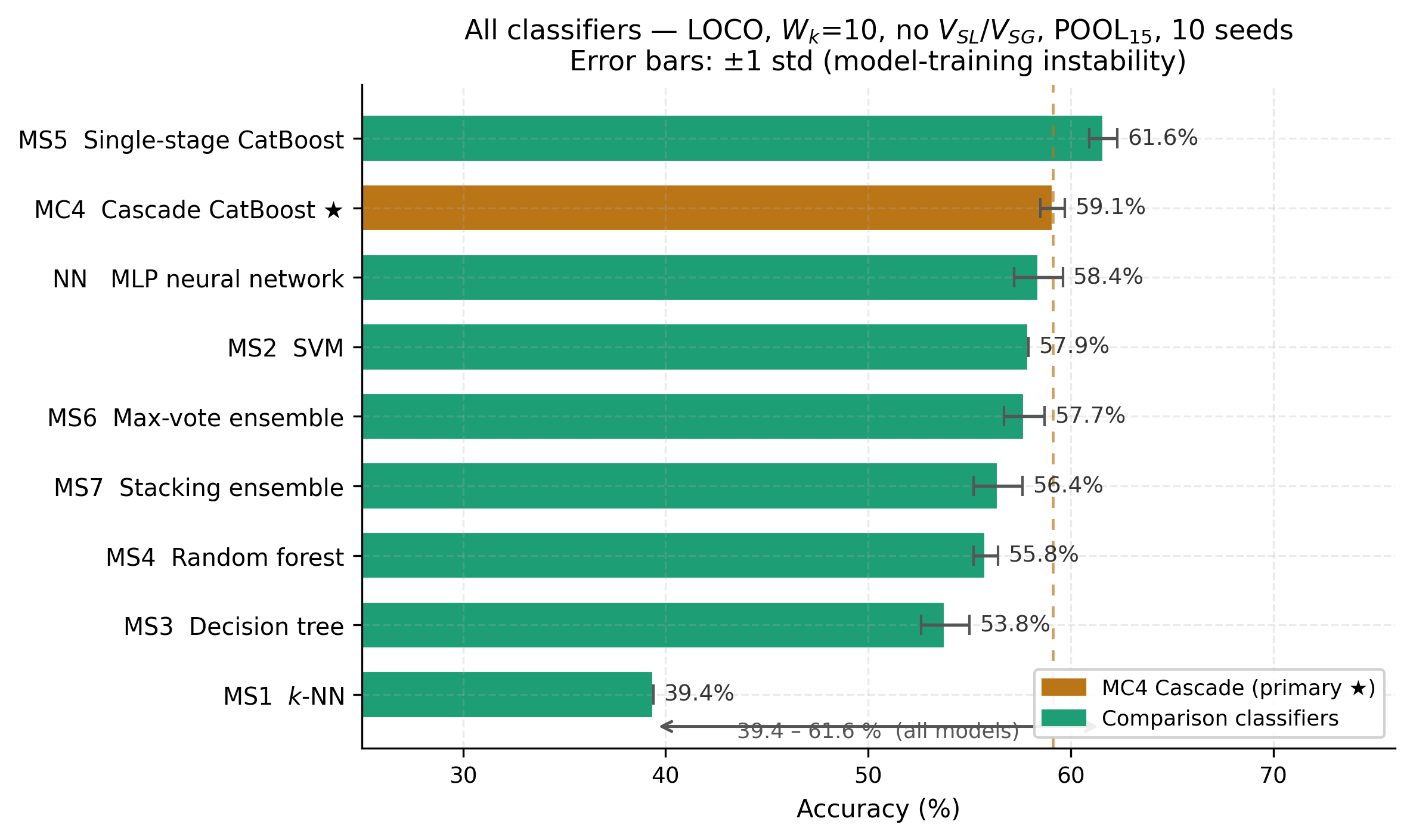}
  \caption{Accuracy of all nine classifiers at $W_k=10$, LOCO, no
  $V_{SL}$/$V_{SG}$, POOL\textsubscript{15}, 10 seeds. The range
  39.4--61.6\% confirms the within-condition versus unseen-condition
  accuracy gap is not specific to the MC4 cascade architecture.}
  \label{fig:fig3}
\end{figure}

Table~\ref{tab:table3} and Fig.~\ref{fig:fig3} report accuracy for all
nine classifiers at $W_k=10$, LOCO, no $V_{SL}$/$V_{SG}$,
POOL\textsubscript{15}, 10 seeds each.

\begin{table}[htbp]
\caption{All classifiers at $W_k=10$, LOCO, no $V_{SL}$/$V_{SG}$,
POOL\textsubscript{15}, 10 seeds. No model exceeds 61.6\% accuracy;
the finding is not specific to any architecture. MS1 (k-NN) and MS2
(SVM) are deterministic, hence $\pm0.0$\% across seeds.}
\label{tab:table3}
\centering
\small
\begin{tabular}{llcccc}
\toprule
\multicolumn{2}{l}{Model} & Accuracy (\%) & Macro F1 & FAR (\%) & MDR (\%) \\
\midrule
MC4 & Cascade CatBoost (primary)$^{\star}$
  & $59.1\pm0.6$ & $0.592\pm0.006$ & $34.1\pm1.0$ & $10.2\pm0.4$ \\
\midrule
MS5 & Single-stage CatBoost & $61.6\pm0.7$ & $0.617\pm0.008$ & $27.9\pm0.6$ & $10.3\pm0.9$ \\
MS2 & SVM                   & $57.9\pm0.0$ & $0.583\pm0.000$ & $29.4\pm0.0$ & $10.3\pm0.0$ \\
MS6 & Max-vote ensemble     & $57.7\pm1.0$ & $0.581\pm0.009$ & $29.0\pm0.8$ & $9.8\pm0.5$  \\
MS7 & Stacking ensemble     & $56.4\pm1.2$ & $0.564\pm0.013$ & $35.1\pm1.2$ & $15.8\pm1.9$ \\
MS4 & Random forest         & $55.8\pm0.6$ & $0.558\pm0.006$ & $33.0\pm0.8$ & $14.4\pm0.4$ \\
MS3 & Decision tree         & $53.8\pm1.2$ & $0.540\pm0.012$ & $40.6\pm1.0$ & $15.2\pm1.1$ \\
NN  & MLP neural network    & $58.4\pm1.2$ & $0.583\pm0.013$ & $16.4\pm1.9$ & $8.0\pm1.0$  \\
MS1 & $k$-NN                & $39.4\pm0.0$ & $0.398\pm0.000$ & $57.4\pm0.0$ & $17.7\pm0.0$ \\
\bottomrule
\end{tabular}\\[4pt]
\footnotesize $^{\star}$MC4 is the pre-specified primary model from~\cite{alquennah2026}.
\end{table}

\subsection{Window-Size Ablation}
\label{sec:windowsize}

\begin{table}[htbp]
\caption{Window-size ablation: Condition~C, 15 conditions, 10 seeds.
Accuracy does not increase monotonically with $W_k$, in contrast to the
within-condition results in~\cite{alquennah2026}.}
\label{tab:table4}
\centering
\begin{tabular}{lccccc}
\toprule
$W_k$ & Duration & Accuracy & Macro F1 & FAR & MDR \\
\midrule
6  & 3.0\,s & $55.4\pm0.7$\% & $55.3\pm0.7$\% & $37.7\pm0.5$\% & $10.1\pm0.2$\% \\
8  & 4.0\,s & $56.0\pm0.8$\% & $56.1\pm0.7$\% & $35.3\pm1.1$\% & $9.5\pm0.1$\%  \\
10 & 5.0\,s & $59.1\pm0.6$\% & $59.2\pm0.6$\% & $34.1\pm1.0$\% & $10.2\pm0.4$\% \\
\bottomrule
\end{tabular}
\end{table}

Larger windows capture more of the slug dynamics within each 30-second
recording but reduce the number of windows extractable per recording,
shrinking the training set under LOCO. Whether the $W_k=10$ advantage
over $W_k=6$ represents a stable property of the flow physics or a
finite-sample artefact cannot be determined from 15 conditions alone.

\subsection{Corrected Noise Robustness}
\label{sec:noiseresults}

\begin{table}[htbp]
\caption{Noise robustness under symmetric contamination: $W_k=10$, LOCO,
no $V_{SL}$/$V_{SG}$, 10 seeds. Accuracy is essentially invariant across
the tested range (58.5--59.1\%); the performance floor is set by the
evaluation protocol, not sensor noise at these FSO levels. The 0\%~FSO
entry matches Condition~C in Table~\ref{tab:table1}.}
\label{tab:table5}
\centering
\begin{tabular}{lcccc}
\toprule
Noise (\%FSO) & Paper (asymm.) & Accuracy (\%) & FAR (\%) & MDR (\%) \\
\midrule
0.0 & 100\% & $59.1\pm0.6$ & $34.1\pm1.0$ & $10.2\pm0.4$ \\
0.1 & 100\% & $58.5\pm0.8$ & $31.9\pm1.0$ & $10.6\pm0.5$ \\
0.2 & 99\%  & $58.9\pm1.1$ & $31.1\pm1.7$ & $10.5\pm0.4$ \\
0.5 & N/A   & $59.0\pm0.7$ & $30.4\pm0.9$ & $10.5\pm0.3$ \\
1.0 & N/A   & $59.0\pm0.4$ & $29.8\pm0.8$ & $10.1\pm0.4$ \\
\bottomrule
\end{tabular}
\end{table}

\begin{figure}[htbp]
  \centering
  \includegraphics[width=\linewidth]{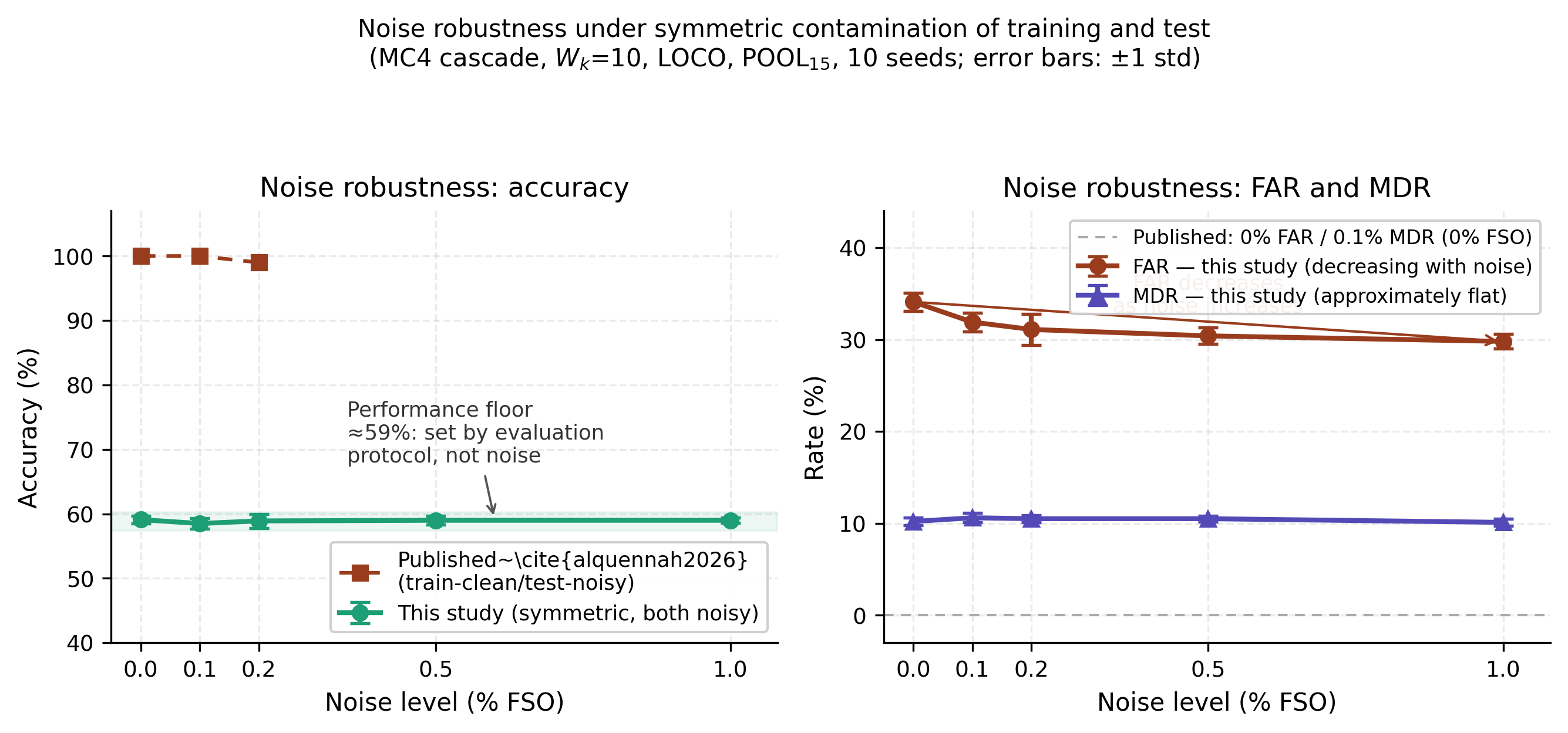}
  \caption{Noise robustness under the corrected symmetric protocol: accuracy
  (left) and FAR/MDR (right) versus noise level. Accuracy remains within
  0.6~pp of the noise-free baseline across all five FSO levels, confirming
  the performance floor is set by the condition-blocked evaluation, not by
  additive sensor noise. FAR decreases from 34.1\% to 29.8\% as noise
  increases; MDR is approximately constant at $\sim$10.3\%.}
  \label{fig:fig4}
\end{figure}

\subsection{Feature Importance}
\label{sec:fimp}

\begin{table}[htbp]
\caption{Top-10 CatBoost feature importances at $W_k=10$, LOCO, no
$V_{SL}$/$V_{SG}$. P2\_max (downstream sensor maximum) dominates
at 41.2\%.}
\label{tab:table7}
\centering
\begin{tabular}{clc}
\toprule
Rank & Feature & Importance (\%) \\
\midrule
1  & P2\_max      & 41.19 \\
2  & P1\_min      &  8.12 \\
3  & P2\_range    &  7.81 \\
4  & P2\_meandiff &  6.38 \\
5  & P1\_max      &  6.20 \\
6  & P2\_std      &  4.91 \\
7  & P1\_rms      &  4.39 \\
8  & P2\_rms      &  4.33 \\
9  & P2\_mean     &  3.71 \\
10 & P1\_mean     &  3.23 \\
\bottomrule
\end{tabular}
\end{table}

\begin{figure}[htbp]
  \centering
  \includegraphics[width=\linewidth]{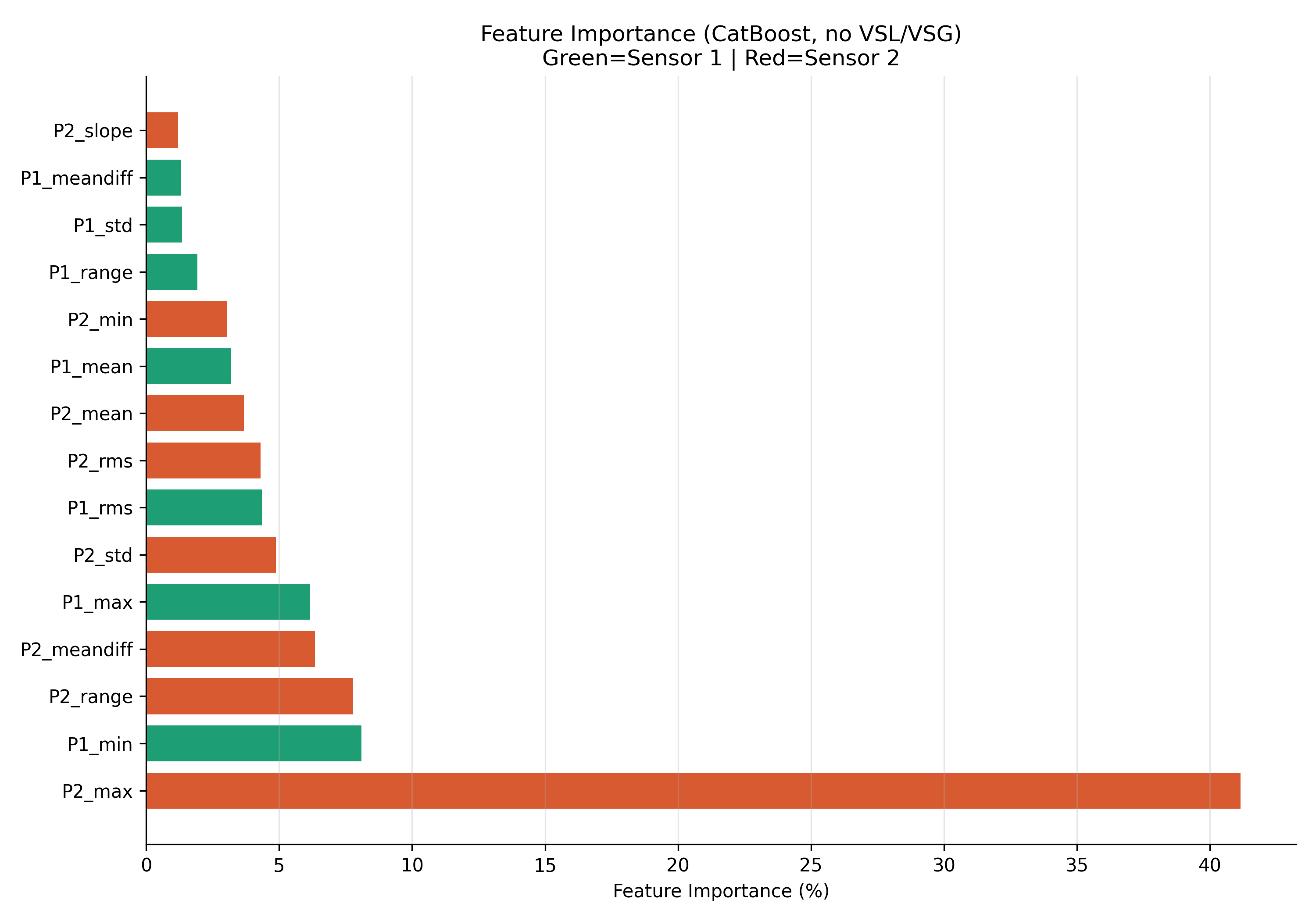}
  \caption{CatBoost feature importances at $W_k=10$, LOCO, no
  $V_{SL}$/$V_{SG}$. Green: upstream sensor S1 features (P1); red:
  downstream sensor S2 features (P2). P2\_max alone accounts for
  41.2\% of total importance.}
  \label{fig:fig5}
\end{figure}

\subsection{External Evaluation: GPLA-12}
\label{sec:external}

To assess whether the accuracy gap under grouped evaluation is specific
to this apparatus or reflects a general pattern, the A/B/C protocol
(Section~\ref{sec:decomp}) was applied to GPLA-12~\cite{li2021gpla},
an independent publicly available acoustic dataset: two microphone
positions, three gas pressure levels (0.2, 0.4, 0.5~MPa), and two noise
conditions, yielding 684 observations across twelve combinations. The
grouping variable is the (microphone, noise) pair; the target is gas
pressure level; and the grouping unit is the domain, defined by four
microphone--noise combinations.

Table~\ref{tab:external} shows that random-split accuracy of
96.8\%$\pm$1.3\% decreases to 33.7\%$\pm$1.1\% under leave-one-group-out
evaluation --- approaching three-class chance level (33.3\%), with a
decrease of 61.4 percentage points. Note that only four domain groups
are available, so group-level uncertainty is substantial; 684 acoustic
segments do not represent 684 independent domain-transfer experiments.
Accuracy near chance level under grouped evaluation suggests limited
transfer to an unseen microphone--noise domain under the evaluated
feature representation, consistent with the rig-based finding across
a different physical system and sensor modality.

\begin{table}[htbp]
\caption{External evaluation on GPLA-12 (ten seeds, four
leave-one-group-out folds). Only four domain groups provide
cross-domain evidence; group-level uncertainty is substantial.}
\label{tab:external}
\centering
\begin{tabular}{p{5.2cm}p{2.8cm}}
\toprule
Condition & Accuracy (\%) \\
\midrule
A: random split, with domain features & $96.8\pm1.3$ \\
B: LOGO (4 groups), with features     & $35.4\pm2.5$ \\
C: LOGO (4 groups), no features       & $33.7\pm1.1$ \\
\midrule
Gap (A$-$B) & $-61.4$\,pp \\
Feature-encoding gap (B$-$C) & $-1.7$\,pp \\
\bottomrule
\end{tabular}
\end{table}

\section{Discussion}
\label{sec:discussion}

\subsection{Reading the $2\times2$ Table}

The interaction contrast $I = A - B - D + C$ (Eq.~\ref{eq:interact})
measures whether the two corrections combine linearly. At $W_k=10$,
$I = 93.3 - 62.6 - 88.4 + 59.1 = 1.4$~pp --- small relative to the
dominant split-protocol component (30.7~pp) and the feature-encoding
component (3.5~pp). At $W_k=6$ and $W_k=8$ the values are $-2.2$~pp
and $-2.7$~pp respectively. All are within the seed-to-seed variability,
supporting approximate additivity.

The practical consequence is straightforward: because the two corrections
interact only weakly, addressing either one in isolation recovers roughly
its share of the accuracy gap. Correcting the split protocol alone
(A$\to$B) recovers 30.7~pp; correcting the feature encoding alone
(A$\to$D) recovers 4.9~pp. Correcting both recovers 34.2~pp. A future
study that fixes only the split protocol will achieve most of the gain;
one that fixes only the feature encoding will not. This asymmetry --- the
split protocol driving roughly 90\% of the gap while feature encoding
drives roughly 10\% --- is consistent across all three window sizes.

The feature-encoding component being larger at shorter windows (6.7~pp
at $W_k=6$ versus 3.5~pp at $W_k=10$) is physically interpretable.
Shorter windows extract less information from the slug dynamics within
each recording, so the explicit $V_{SL}$/$V_{SG}$ coordinates carry
relatively more discriminative weight at small window sizes.

\subsection{Detection Performance at Unseen Conditions}

The per-class results in Table~\ref{tab:table2} reveal a performance
asymmetry that survives the protocol corrections. The no-leak class
achieves F1~=~70.8\%, while two-point/multiple leakage achieves 50.3\%.
Relative to count discrimination, binary detection --- telling the
operator that a leak is present at all --- retains more discriminative
capacity at unseen operating conditions. This has a physical basis:
binary detection asks whether the pressure dynamics between S1 and S2
have been disturbed, and any leak at any location creates a detectable
pressure perturbation. Distinguishing one 1.8~mm orifice from two
orifices of different sizes is a harder problem at 2-Hz temporal
resolution, where differences may lie in wave arrival timing or amplitude
ratios that a 0.5-second sample cannot resolve.

Both capabilities nonetheless fall substantially below the originally
reported values. The false alarm rate of 34.1\% at unseen conditions
means more than one in three no-leak windows would trigger a field alarm;
the missed detection rate of 10.2\% means roughly one in ten genuine
leak events would go undetected. These are laboratory estimates from a
single six-metre rig at a specific range of air--water velocities. They
cannot be transferred directly to any operational deployment without a
prospectively designed field study using conditions drawn independently
from the intended operating envelope.

The neural network baseline (Table~\ref{tab:table3}) achieves a false
alarm rate of 16.4\% --- roughly half the cascade value --- despite
similar overall accuracy (58.4\%). This suggests that architecture choice
affects operational alarm behaviour even when aggregate accuracy is
comparable, and warrants dedicated investigation in follow-on work.

\subsection{Operating Conditions at Chance Level}

Per-fold analysis shows that conditions at $V_{SL}=2.21$~m/s with
$V_{SG}\in\{0.12, 0.19\}$~m/s return accuracy of exactly 33.3\% ---
three-class chance --- across all ten seeds. Under this model and
feature representation, the 2-Hz dynamic pressure signal at those
conditions did not provide transferable discriminative information about
the leak state from the training conditions. Whether this reflects a
physical limit of low-frequency pressure sensing at high liquid velocities,
a shortage of diversity among the training conditions, or a combination
of the two, remains an open question. Addressing it requires higher
sampling rates, additional sensors, or a richer set of training conditions
at high $V_{SL}$.

\subsection{On the Use of $V_{SL}$ and $V_{SG}$ as Model Inputs}

Encoding operating conditions directly as features is not inherently
a methodological error. If the deployed system receives live flow-meter
readings or set-point signals, including them as covariates is appropriate.
The feature-ablation test (A$\to$D, encoding corrected while split
unchanged) recovers only 4.9~pp at $W_k=10$, far less than the split
correction alone (30.7~pp). The large accuracy in the original
study~\cite{alquennah2026} therefore came primarily from within-condition
evaluation, not from the operating-condition features.

\subsection{External Evaluation in Context}

The 61.4-percentage-point decrease observed for GPLA-12 under grouped evaluation
is consistent with the accuracy decrease obtained on the six-metre horizontal air--water flow loop. However, four domain groups provide
limited statistical power at the group level --- the $\pm2.5$~pp variance
under the LOGO condition with features reflects genuine group-level
uncertainty. The result does not constitute independent statistical
confirmation; it does show that a model trained on one recording setup
failed at a different recording setup for the same target variable, a
pattern seen in both experiments.

\subsection{Diagnostic Checklist for Future Studies}

Six checks are derived directly from the protocol characteristics
identified in Section~\ref{sec:issues}. They apply to any ML study on
grouped experimental multiphase-flow data.

\begin{enumerate}
\item Withhold each test condition from training; report results under
LOCO or an equivalent condition-blocked protocol.
\item Construct windowed features independently within each fold; never
apply feature extraction to the full dataset before splitting.
\item Report performance with and without operating-condition features;
note whether those features would be available to the deployed system.
\item Verify that unlabelled data columns are genuine independent
recordings before treating them as additional conditions.
\item Apply noise to both training and test when evaluating robustness
under the assumption that noise is present at both collection and inference.
\item Report results under all ablation conditions, not only the
best-performing configuration.
\end{enumerate}

\subsection{Directions for Future Work}
\label{sec:futurework}

Several directions follow from the present results. The most immediate is
expanding the condition space, particularly at $V_{SL}=2.21$~m/s where
every architecture tested returns chance-level accuracy at unseen conditions.
Whether higher sampling rates or physics-informed features that capture
the spatial pressure gradient between S1 and S2 can improve transfer at
those conditions is testable and has direct implications for system design.

The symmetric noise protocol models one plausible field scenario. Correlated
process noise --- slug-induced pressure transients intrinsic to horizontal
two-phase flow --- may interact with leak signatures in ways that Gaussian
noise cannot capture. Extending the protocol to cover asymmetric and
correlated noise scenarios would sharpen the picture.

The window-level inner cross-validation introduces a residual within-recording
dependence. A recording-block inner split would eliminate it and would allow
a cleaner comparison of hyperparameter strategies without the confound.

Finally, the FAR and MDR reported here are laboratory estimates. A
prospectively designed field study with conditions drawn from the intended
operating envelope and an agreed alarm threshold is the minimum required
before any deployment decision.

\section{Literature Survey of Leak Detection Contributions}
\label{sec:literature}

Table~\ref{tab:literature} situates this work within published research
on leak detection in multiphase pipeline systems. Studies marked $\dagger$
are the subject of the evaluation audit conducted here.


{\footnotesize
\setlength{\tabcolsep}{3pt}
\begin{longtable}{%
  >{\raggedright\arraybackslash}p{2.7cm}
  >{\raggedright\arraybackslash}p{4.8cm}
  >{\raggedright\arraybackslash}p{2.8cm}
  >{\raggedright\arraybackslash}p{2.0cm}
  >{\raggedright\arraybackslash}p{2.1cm}}
\caption{Contributions to leak detection research in multiphase pipeline
systems. $\dagger$ marks studies subject to the evaluation-protocol
audit in this paper.}
\label{tab:literature}\\
\toprule
\textbf{Authors} & \textbf{Method / Parameters} &
\textbf{Configuration} & \textbf{$d$\,/\,$L$} & \textbf{Leak} \\
\midrule
\endfirsthead
\multicolumn{5}{l}{\small\textit{Table~\ref{tab:literature} continued.}}\\
\toprule
\textbf{Authors} & \textbf{Method / Parameters} &
\textbf{Configuration} & \textbf{$d$\,/\,$L$} & \textbf{Leak} \\
\midrule
\endhead
\midrule\multicolumn{5}{r}{\small\textit{Continued.}}\\\endfoot
\bottomrule\endlastfoot

\multicolumn{5}{l}{\textbf{Experimental modelling}} \\[1pt]

Siebenaler et al.\ (2017)
  & Gamma densitometer, hydrophones, high-speed camera
  & Horiz.\ slug, air--water
  & 127\,mm / 6\,m & 3.2--9.5\,mm \\[2pt]

Stajanca et al.\ (2018)
  & Fibre optic, acoustics, accelerometer
  & Horiz.\ air
  & 114\,mm / 38\,m & 1--8\,mm \\[2pt]

Ji et al.\ (2018)
  & Acoustics, dynamic pressure
  & Horiz.\ strat./wavy/slug
  & 80\,mm / 14.3\,m & 3--6\,mm \\[2pt]

Li et al.\ (2022)
  & Dynamic pressure, acoustics
  & Horiz.\ stratified
  & 40\,mm / 48\,m & 8\,mm \\[2pt]

Meng et al.\ (2024)
  & Ultrasonic, pressure, mass flow
  & Horiz.\ strat./slug/annular
  & 32\,mm / 4.8\,m & 3\,mm \\[2pt]

Khan et al.\ (2024)~\cite{khan2025}
  & ERT (void fraction)
  & Horiz.\ plug/bubbly
  & 50.8\,mm / 6\,m & 1.8--3\,mm \\[2pt]

Li and Yao (2021)~\cite{li2021gpla}
  & Acoustic emission; 12 mic./noise combos; external evaluation here
  & Horiz.\ gas pipeline
  & --- & 1 pt \\[2pt]

\multicolumn{5}{l}{\textbf{Numerical modelling}} \\[1pt]

Adegboye et al.\ (2021)
  & Pressure, void fraction (CFD)
  & Horiz.\ stratified
  & 60\,mm / 18\,m & 1.5--18\,mm \\[2pt]

Vandrangi et al.\ (2021)
  & Pressure, temp., volume (OLGA)
  & Horiz.\ two-phase
  & 203\,mm / 100\,m & --- \\[2pt]

Lin et al.\ (2024)
  & Pressure, void fraction
  & Subsea jumper
  & 48\,mm / 3.6\,m & 15\,mm \\[2pt]

Ren et al.\ (2024)
  & Void fraction, volume flow
  & Horiz.\ plug
  & 50\,mm / 6\,m & 15\,mm \\[2pt]

Ferroudji et al.\ (2024)~\cite{ferroudji2024b}
  & Dynamic pressure, hydrophone
  & Horiz.\ plug/bubbly
  & 50.8\,mm / 3\,m & 1.8--3\,mm \\[2pt]

\multicolumn{5}{l}{\textbf{Machine learning and other approaches}} \\[1pt]

Santoso et al.\ (2014)
  & ANN (pressure drop)
  & Horiz.\ plug
  & 24\,mm / --- & --- \\[2pt]

Zhang et al.\ (2023)
  & MFCC-LSTM
  & Horiz.\ strat./slug/annular
  & 32\,mm / 1.2\,m & 2--5\,mm \\[2pt]

Al-Ammari and Sleiti (2024)
  & ANN (temp., pressure, mass flow)
  & Horiz.\ multiphase
  & 320\,mm / 13\,km & 30\,mm \\[2pt]

Barooah et al.\ (2024)~\cite{barooah2024}
  & Mechanistic + data-driven hybrid
  & Horiz.\ multiphase
  & 76--172\,mm / 0.6--6.5\,km & 12--76\,mm \\[2pt]

Ferroudji et al.\ (2025)$^\dagger$~\cite{ferroudji2025}
  & Dyn.\ pressure; k-NN, SVM, RF, stacking; unspecified CV\@.
    ML5: 81.8\% reported (our LOCO: 62.6\%).
    ML6: 89.5\% reported; not replicable---DP/NF1 sensors absent
  & Horiz.\ air--water
  & 50.8\,mm / 6\,m & 1.8--3\,mm \\[2pt]

Alquennah et al.\ (2026)$^\dagger$~\cite{alquennah2026}
  & Dyn.\ pressure; cascaded CatBoost; time-based split;
    reported 100\%
  & Horiz.\ air--water
  & 50.8\,mm / 6\,m & 1.8--3\,mm \\[2pt]

\textbf{This study}
  & Dyn.\ pressure; LOCO; $2\!\times\!2$ ablation; 34\,pp gap;
    $|I|\!\leq\!3$\,pp; accuracy 50.9--59.1\%
  & Same as $\dagger$
  & 50.8\,mm / 6\,m & 1.8--3\,mm \\

\end{longtable}
}

\section{Data and Code Availability}
\label{sec:datacode}

The analysis code used for data loading, feature extraction, model training,
evaluation, and generation of the reported tables and figures is available in
a public GitHub repository at
\url{https://github.com/issahnazif-afk/100-percent-warning-sign}. The exact
version used for this study is archived at Zenodo under DOI
\url{https://doi.org/10.5281/zenodo.24218226}. The final numerical results and
figures are reported in this article; the repository contains the scripts
required to regenerate them from the source data. The raw sensor recordings are
not publicly available at this time. Researchers wishing to request access to
the recordings may contact the corresponding author.

\section{Conclusion}
\label{sec:conclusion}

Five findings follow from this study.

\textbf{1.}~A pressure-based ML leak-detection model reporting 100\%
accuracy under within-condition evaluation achieves 50.9\%$\pm$0.5\% on
the nine-condition subset matching the original study design and
59.1\%$\pm$0.6\% on the full 15-condition dataset under condition-blocked
evaluation. The same conditions, the same raw observations, the same
sensors: the only change is whether the model encounters the test condition
during training.

\textbf{2.}~The within-condition versus unseen-condition accuracy gap of
34.2~pp at $W_k=10$ decomposes into a split-protocol component
(30.7~pp) and a feature-encoding component (3.5~pp). The interaction
contrast at $W_k=10$ is $+1.4$~pp ($-2.2$~pp at $W_k=6$;
$-2.7$~pp at $W_k=8$), supporting approximate additivity: each
correction contributes its share approximately independently.

\textbf{3.}~Under condition-blocked evaluation at $W_k=10$, the false
alarm rate rises from 0.0\% to 34.1\%$\pm$1.0\% and the missed detection
rate from 0.1\% to 10.2\%$\pm$0.4\%. Binary leak detection achieves
F1~=~70.8\% while two-point leakage discrimination achieves 50.3\%.
Both fall substantially below the originally reported values, and the
laboratory FAR and MDR should not be treated as deployment-ready
performance guarantees without a prospective field study.

\textbf{4.}~Conditions at $V_{SL}=2.21$~m/s with $V_{SG}\in\{0.12,
0.19\}$~m/s return chance-level accuracy (33.3\%) across all ten seeds
under every model tested. Under this feature representation and model
configuration, the 2-Hz dynamic pressure signal at those conditions did
not provide transferable discriminative information about the leak state.
This is a physically interpretable result with direct implications for
the design of future sensing strategies at high liquid superficial
velocities.

\textbf{5.}~On the independent GPLA-12 acoustic dataset --- a different
rig, sensor modality, and target variable --- random-split accuracy of
96.8\% decreases to 33.7\%$\pm$1.1\% under grouped evaluation, a
61-percentage-point decrease approaching chance level. The within-condition
versus unseen-condition accuracy gap is not specific to this flow rig.

The corrected accuracy of 59.1\%$\pm$0.6\% is a starting point, not a
ceiling. Additional experimental conditions --- especially at
$V_{SL}=2.21$~m/s --- and higher-resolution sensing are the most direct
paths to improved cross-condition generalisation. The evaluation protocol
and diagnostic checklist provided here are designed to make future
progress on that goal reproducible and comparable across studies.

\section*{CRediT authorship contribution statement}

\textbf{Issah N. Suleiman:} Conceptualization, Methodology, Software,
Formal analysis, Investigation, Visualization, Writing -- original draft.
\textbf{Sormeh Serpoosh:} Methodology, Validation, Writing -- review and
editing. \textbf{Nadine Elkholy:} Investigation, Data curation, Validation,
Writing -- review and editing. \textbf{Hicham Ferroudji:} Investigation,
Resources, Writing -- review and editing. \textbf{Mohammad Azizur Rahman:}
Supervision, Funding acquisition, Resources, Project administration, Writing
-- review and editing. \textbf{Matthew Hamilton:} Conceptualization,
Methodology, Supervision, Project administration, Writing -- review and
editing.

\section*{Declaration of competing interest}

The authors declare that they have no known competing financial interests or
personal relationships that could have appeared to influence the work
reported in this paper.

\section*{Acknowledgment}

This publication was made possible by the grant NPRP14S-0321-210080 from the Qatar National Research Fund (a member of the Qatar Foundation). The authors would like to acknowledge QNRF for providing support and guidance, without which this work would not have been possible. The authors also acknowledge support from the Qatar Research, Development and Innovation (QRDI) Council through the Climate Change and Environment Call (CCEC), project CCEC02-0216-250065. This work was also partially funded by the Canada First Research Excellence Fund (CFREF) through the Transforming Climate Action program. Statements made herein are solely the responsibility of the authors.

\end{document}